\documentclass{elsarticle}
\usepackage[utf8]{inputenc}

\usepackage{natbib}
\usepackage{graphicx}
\usepackage{bm}
\usepackage{amsmath}
\usepackage{mathtools}
\usepackage{amsfonts}
\usepackage{amssymb}
\usepackage{hyperref}

\begin{document}

\begin{frontmatter}

\title{Probabilistic Solar Power Forecasting: Long Short-Term Memory Network vs Simpler Approaches}

\author[myaddress]{Vinayak Sharma \corref{mycorrespondingauthor}}
\cortext[mycorrespondingauthor]{Corresponding author}
\ead{vsharm12@uncc.edu}
\author[myaddress2,myaddress3]{Jorge \'Angel Gonz\'alez Ordiano}
\author[myaddress2]{Ralf Mikut}
\author[myaddress5]{Umit Cali}
%
\address[myaddress]{University of North Carolina at Charlotte}
\address[myaddress2]{Institute for Automation and Applied Informatics, \\ Karlsruhe Institute of Technology}
\address[myaddress3]{Colorado State University}
\address[myaddress5]{Norwegian University of Science and Technology}

\begin{abstract}
The high penetration of volatile renewable energy sources such as solar, make methods for coping with the uncertainty associated with them of paramount importance. Probabilistic forecasts are an example of these methods, as they assist energy planners in their decision-making process by providing them with information about the uncertainty of future power generation. Currently, there is a trend towards the use of deep learning probabilistic forecasting methods. However, the point at which the more complex deep learning methods should be preferred over more simple approaches is not yet clear. Therefore, the current article presents a simple comparison between a long short-term memory neural network and other more simple approaches. The comparison consists of training and comparing models able to provide one-day-ahead probabilistic forecasts for a solar power system. Moreover, the current paper makes use of an open source dataset provided during the Global Energy Forecasting Competition of 2014 (GEFCom14).


\end{abstract}


\end{frontmatter}

\section{Introduction}

Over the past couple of years solar power has become one of the most popular renewable energy sources (RES). Unfortunately, the generation of solar power depends completely on the Sun~\cite{BAK2002991}. This dependency on weather adds uncertainty and variability to the generation of solar power. To deal with this uncertainty, solar forecasts are made in-order to predict the future power generation. 

Solar power forecasts can be categorized into deterministic and probabilistic forecasts~\cite{Antonanzas16}. Some examples of deterministic forecasting methods present in literature can be found in~\cite{Abuella7, DIAGNE201365, doi:10.1002/widm.1235, Vinayak, Sharma}. While deterministic forecasts predict only the expected future generation, probabilistic forecasts offer a description of the forecast uncertainty. This additional information helps in managing resources, as well as, in calculating risks associated with future decisions~\cite{Scheduling, proba_over}. Furthermore, economic benefits can also be gained from using probabilistic forecasts, since they improve the decision making capabilities within electricity markets~\cite{ROULSTON2003585}.

Various methodologies to generate probabilistic solar power forecasts have been discussed in literature. For example, nearest neighbor approaches~\cite{7285696}, vector auto-regressive (VAR) models~\cite{BESSA201516}, methods for estimating volatility~\cite{fliess:hal-01736518}, and ensemble models~\cite{ALESSANDRINI201595}. Additionally, examples of solar power probabilistic forecasting using  deep learning techniques can also be found in literature, e.g., in~\cite{7844673}. However, even though deep learning methodologies have gained in popularity in the past couple of years, they have often under-performed in terms of accuracy when compared to other statistical forecasting techniques~\cite{10.1371/journal.pone.0194889}. 

For this reason, the current article presents a small experiment with the goal of defining a starting point for understanding the limitations of deep learning probabilistic forecasting methodologies. To be more specific, the experiment consists in training, evaluating, and comparing solar power probabilistic forecasts based on quantile regressions~\cite{Fahrmeir13} obtained using a long-short term memory (LSTM) neural network (i.e. a deep learning approach) and more simple techniques (i.e. polynomials and a fully connected artificial neural network). Furthermore, the open source dataset of the Global Energy Forecasting Competition of 2014 is used for the experiment.

The remainder of the current paper is divided as follows. Section~\ref{sec_methods} presents a brief description of the various methods tested. Thereafter, Section~\ref{sec_experiment} describes more in detail the conducted experiment. Afterwards, Section~\ref{sec_results} presents the obtained results and finally, Section~\ref{sec_conclusion} offers the conclusion and outlook of this work.



\section{Methods}\label{sec_methods}

Quantile regressions are useful at estimating the uncertainty of a time series' future. A finite time series $\{P[k];k = 1,\dots,K\}$ is defined as a sequence of observations $P[k]$ measured at different points in time; with the timestep $k$ defining the order of the observation in the sequence and $K\in \mathbb{N}_{>0}$ representing the time series' length. 

In turn, a quantile regression can be viewed as a model able to estimate a quantile with a probability $\tau\in(0,1)$ of a future value $P[k+H]$ at a forecast horizon $H\in \mathbb{N}_{>0}$. For instance, a quantile regression that takes auto-regressive and exogenous values as input can be defined as:
\begin{equation}
\hat{P}_{\tau}[k+H] = f(P[k],\dots,P[k-H_{\mathrm{l}}],\mathbf{u}^{T}[k],\dots,\mathbf{u}^{T}[k-H_{\mathrm{l}}];\hat{\boldsymbol{\theta}}_{\tau})\text{ ;}
\end{equation}
where $\hat{P}_{\tau}[k+H]$ is the quantile estimate, $H_{\mathrm{l}}$ is the number of used lags, and $\mathbf{u}[k]$ represents a vector containing observations of exogenous time series at timestep $k$. Moreover, $\hat{\boldsymbol{\theta}}_{\tau}$ is a vector containing the estimated regression parameters, which are traditionally obtained through the minimization of the sum of pinball-losses~\cite{Fahrmeir13}. Furthermore, one of the most important properties of quantile regressions is the fact that pairs of them can be combined to form intervals with a certain probability of containing a future time series' value (i.e. probabilistic forecasts). Finally, more detailed information of the models used in the present article can be found in the following sections. 

\subsection{Simple Models}

\subsubsection{Polynomials}

Quantile regressions trained using a polynomial model are multiple linear quantile regressions, whose features can be raised to a maximal allowed degree.
Some examples of this type of model can be found in~\cite{GonzalezOrdiano17}.

\subsubsection{Fully Connected Artificial Neural Network}

The fully connected artificial neural network (FCANN) used in the present article is a simple multilayer perceptron~\cite{Hastie16} with only one hidden layer. The advantage of this model over the polynomials is the fact that it can more easily describe non-linear relations between its inputs and its desired output (i.e. the solar power time series' future values). It needs to be mentioned, that the FCANN quantile regressions are trained using a method described in~\cite{GonzalezOrdiano16a}.

\subsection{Long Short-Term Memory Neural Network}

A Long Short-Term Memory (LSTM) neural network model~\cite{DBLP:journals/corr/Lipton15} is part of the Recurrent Neural Network (RNN) family. An RNN is a neural network able to learn temporal dependencies in data. In other words, RNNs can establish a correlation between the previous data points and the current data point in the training sequence~\cite{LSTM-load}. This property makes them ideal for solar power forecasting. However, in cases where long-term relationships need to be learned, traditional RNNs face the problem of gradient vanishing. LSTMs solve this issue by using an additional unit called a memory cell~\cite{DBLP:journals/corr/Lipton15} that helps them in learning and explaining long-term relationships~\cite{7844673}. LSTM quantile regressions can be obtained using the pinball-loss as error function during training.

\section{Experiment}\label{sec_experiment}

\subsection{Data}

The dataset used comes from the solar track of the Global Energy Forecasting Competition of 2014 (i.e. GEFCom14)~\cite{Hong16}. It contains three different sets of time series with hourly power measurements of three solar power systems in Australia (normalized to values between 0 and 1), as well as, a number of corresponding weather forecast time series for the period of April $1^{st}$, 2012 to July $1^{st}$, 2014. In the present work, only the forecast weather time series containing forecasts of the solar surface radiation, solar thermal radiation, and top net solar radiation from the 1st zoneID are used. Additionally, the data of only one of the solar power systems is utilized; with 70\% of the data used for training and 30\% for testing.

\subsection{Experiment Description}

The experimental setup, to compare the performance of the LSTM to that of the other models, consists in forecasting daily 99 quantiles (i.e. $\tau = 0.01,0.02,$ $\dots, 0.99$) of the next 24 hours of solar power generation (i.e. $H = 24$, due to the time series' hourly resolution). Furthermore, the same input data is used for all quantile regressions; i.e. the solar power measured over the past 24 hours and the forecast radiation values for the next day.
 
The polynomial models used have maximal allowed degrees of one up to three, hence they are referred to as Poly1, Poly2, and Poly3. In turn, the simple FCANN models are multilayer perceptrons with one hidden layer and 10 hidden neurons with a tanh activation function. Additionally, a forward feature selection is applied to select the four most relevant features with which both the polynomial and FCANN models are later trained. Moreover, to improve the forecast accuracy, the night values are removed during training and automatically set to zero during testing. Note that all polynomial and FCANN models are trained using the MATLAB open-source toolbox SciXMiner~\cite{Mikut17}. 

Finally, an LSTM model with one input layer, one hidden layer, and one output layer is developed using the Keras API~\footnote{\href{https://keras.io/}{keras.io}}. The hidden layer consists of 100 hidden neurons with a sigmoid activation function. Additionally, dropout is applied to avoid overfitting the model. Notice that the neural network architecture (i.e. its hyper-parameters) was selected after conducting a series of preliminary experiments. 

The value used to evaluate the results on a test set of size $N$ is the pinball-loss averaged over all the estimated quantiles (as in~\cite{Hong16}), i.e.:
\begin{equation}\label{eq_pinabllLoss}
\begin{split}
Q_{\mathrm{PL}} & = \operatorname{mean}\{Q_{\mathrm{PL},0.01},\dots,Q_{\mathrm{PL},\tau},\dots,Q_{\mathrm{PL},0.99}\}, \text{with}\\
Q_{\mathrm{PL},\tau} & = \dfrac{1}{N} \sum^{K-H}_{k = 1} \begin{dcases}
(\tau-1)~(\hat{P}_{\tau}[k+H]-P[k+H]) & \text{, if } P[k+H] > P_{\tau}[k+H] \\
\tau~(\hat{P}_{\tau}[k+H]-P[k+H]) & \text{, else} 
\end{dcases}
\end{split} \text{ .}
\end{equation}
In the previous equation, $Q_{\mathrm{PL},\tau}$ is the pinball-loss obtained by a quantile regression with probability $\tau$, while $Q_{\mathrm{PL}}$ is the average of the pinball-losses obtained by all estimated regressions. Please notice that a comparison based on computation time is excluded from the present article, as some models are created with MATLAB and others with Python. Nevertheless, due to its relevance, such a comparison is to be done in future related works.

\section{Results}\label{sec_results}

The results on the test set from the above described experiments are presented in Table~\ref{tab:pinball_loss}. 
\begin{table}[!hbt]
\centering
  \begin{tabular}{ccl}
    \\Model & Avg. Pinball-loss [\%] \\  [0.5ex] 
		\hline 
		Poly1 & 1.70  \\ 
		Poly2 & 1.59 \\
		Poly3 & 1.66 \\ 
	    FCANN  & 1.43 \\ \hline
		LSTM & 1.43 \\ 
\end{tabular}
  \caption{Average pinball-loss test set results from all the models}
    \label{tab:pinball_loss}
\end{table}

As the contents of Table~\ref{tab:pinball_loss} show, the LSTM outperforms all of the polynomial models. Nonetheless, the difference in pinball-loss between the LSTM model and the best performing polynomial model is not significantly large, as it just amounts to $0.15\%$. Additionally, the FCANN model has in average the same performance as the more complex LSTM model. The underwhelming performance of the LSTM regressions may be caused by different reasons. For instance, their extensive need for a large training dataset, as it is known that deep learning methodologies need large amounts of data to accurately learn the relationship between the dependent and the independent variables~\cite{Najafabadi2015}. Furthermore, the manually selected hyper-parameters may also be behind the LSTM's underwhelming performance, as this manual selection does not assure that the optimal set of parameters is found. Another explanation could be that the existing real-world non-linearities can be covered by FCANN as good as by LSTM



\begin{figure}[hbt]
    \centering
    \includegraphics[width=\textwidth]{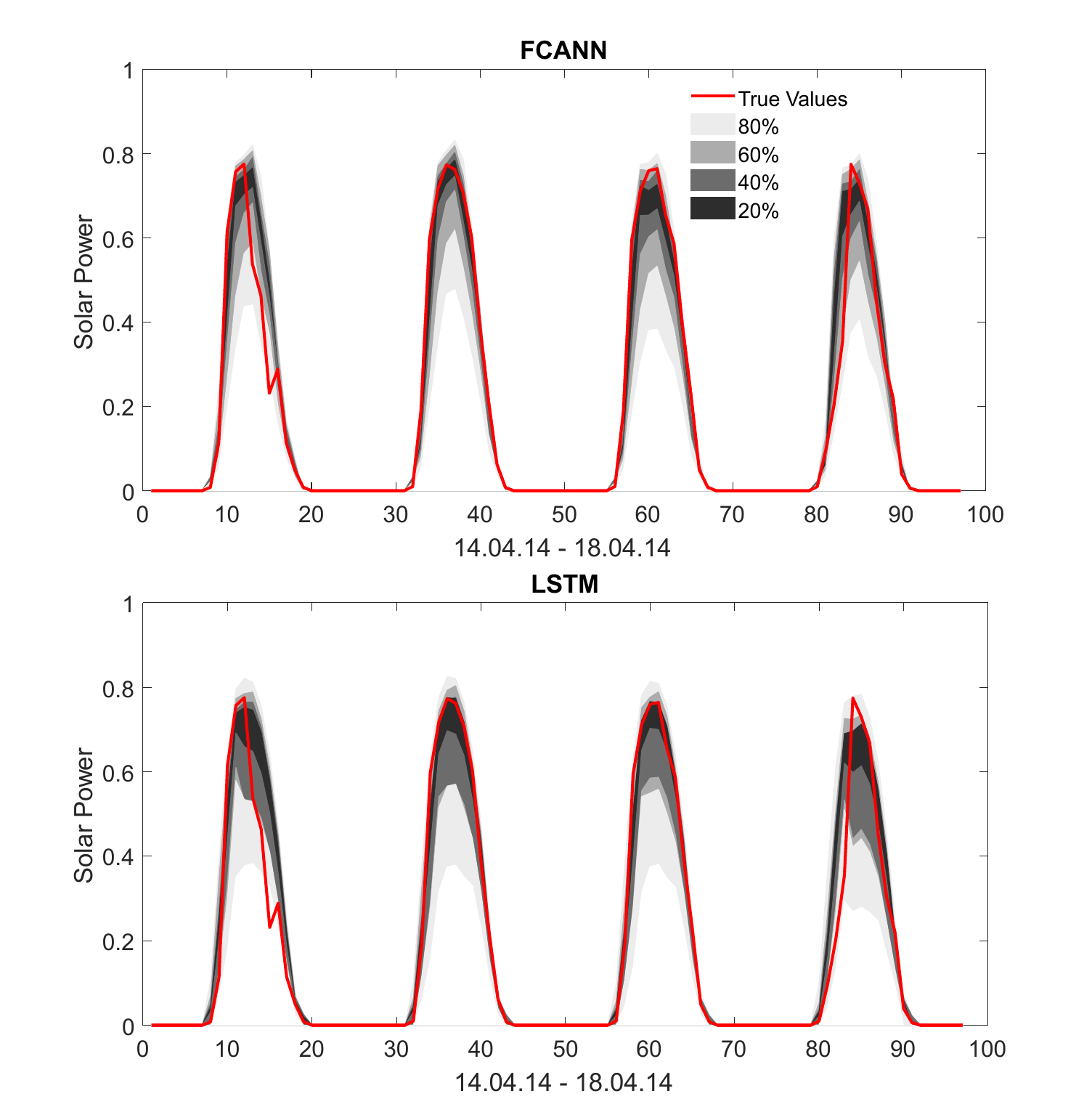}
    \caption{Interval forecasts obtained with the FCANN and LSTM quantile regressions}
    \label{fig:results}
\end{figure}

For the sake of illustration, Figure~\ref{fig:results} depicts the interval forecasts obtained by the FCANN and LSTM models.

As can be seen in Figure~\ref{fig:results}, the LSTM intervals seem to be larger than the ones obtained by the FCANN regressions. Therefore it can be argued, that the LSTM may be overestimating in some degree the uncertainty. This aspect needs to be considered in future related works, if the accuracy of the herein LSTM-based probabilistic forecasts is to be improved.


\section{Conclusion and Outlook}\label{sec_conclusion}

The main contribution of the current article is to present a comparison between a long short-term memory (LSTM) model and other more simple approaches; specifically some polynomial models and a simple fully connected artificial neural network (FCANN). The comparison consists in obtaining and evaluating 24 hour ahead probabilistic solar forecasts. The experiment shows that the LSTM model performs slightly better than the polynomials and obtains the same results as the FCANN. Therefore, it can be argued that the complex LSTM may not always provide the best solution, at least not for the dataset evaluated in this paper. Henceforth, the current article recommends the use of simpler/classical forecasting methodologies as a preliminary benchmarking step before exploring more complex deep learning methods. 

Also, since the underwhelming performance of the LSTM may be caused by a sub-optimal selection of hyper-parameters, hyper-parameter selection via automated machine learning (AutoML) techniques has to be studied in future related works. Moreover, aspects like multiple runs of the neural networks and computation time need also to be taken into consideration in future experiments. At the same time, comparisons as the one presented herein for the case of probabilistic wind and/or load forecasts also need to be studied in the future.

\section*{Acknowledgement}
The present contribution is supported by the Helmholtz Association under the Joint Initiative ‘‘Energy System 2050 — A Contribution of the Research Field Energy’’

\section*{References}
\bibliographystyle{abbrv}
\bibliography{references}

\end{document}